\documentclass[onesided]{article}

\usepackage{PRIMEarxiv}
\usepackage[utf8]{inputenc} 
\usepackage[T1]{fontenc}    
\usepackage{hyperref}       
\usepackage{url}            
\usepackage{booktabs}       
\usepackage{amsfonts}       
\usepackage{nicefrac}       
\usepackage{microtype}      
\usepackage{lipsum}
\usepackage{fancyhdr}       
\usepackage{graphicx}       
\graphicspath{{media/}}     
\usepackage{ulem} 
\hypersetup{hidelinks}
\author{
  Dilip Sarkar\\
  Department of Computer Science \\
  University of Miami \\
  Coral Gables\\
  \texttt{sarkar@miami.edu} \\
  \And
Md. Safayet Islam\\
  Department of Computer Science \\
  University of Miami \\
  Coral Gables\\
  \texttt{mxi451@miami.edu}\\
  \And
  Liang Liang \\
  Department of Computer Science \\
  University of Miami \\
  Coral Gables\\
  \texttt{liang.liang@miami.edu}\\
}

\usepackage{amsmath,amssymb,amsfonts}
\usepackage{url}
\usepackage{graphicx}
\usepackage{textcomp}
\usepackage{xcolor}
\usepackage{tabularx}
\usepackage{adjustbox}
\usepackage{multirow}
\usepackage{array}
\usepackage[utf8]{inputenc} 
\usepackage[T1]{fontenc}    
\usepackage{hyperref}       
\usepackage{url}            
\usepackage{booktabs}       
\usepackage{nicefrac}       
\usepackage{microtype}      
\usepackage{caption, subcaption}
\usepackage{float}
\usepackage{multirow}
\usepackage[table]{xcolor}
\usepackage{arydshln}
\usepackage{paralist}
\usepackage{tabularray}
\usepackage{algorithm}
\usepackage{algpseudocode}
\usepackage{nccmath}
\newtheorem{definition}{Definition}
\newtheorem{assumption}{Assumption}
\newtheorem{lemma}{Lemma}
\def\BibTeX{{\rm B\kern-.05em{\sc i\kern-.025em b}\kern-.08em
    T\kern-.1667em\lower.7ex\hbox{E}\kern-.125emX}}

\begin{document}
\title{Evaluation of MLLM-Agnostic Plug-and-Play Keyframe  Selection  Methods for  Long Video Understanding
}
\maketitle
\begin{abstract}
Multimodal large language models (MLLMs) cannot process every frame of a long video because of limitations in visual-token and computational budgets. Three primary approaches have been proposed to enhance their long-video understanding capabilities: 
\begin{inparaenum}[$(i)$] \item \textit{Retraining an MLLM} on a large video corpus and/or extending its input length; 
\item \textit{Training an adapter} for a specific MLLM that takes the entire video and the query as input and selects the most relevant video frames; and 
\item \textit{Developing a training-free, plug-and-play (PaP) adapter} that is MLLM-agnostic. We refer to the third approach as PaP keyframe selection. A PaP method may use only candidate video frames without considering the query, or it may use both candidate video frames and the query.\end{inparaenum} \par The first approach is prohibitively expensive. The second approach requires substantial training time and computational resources, but it is accessible to many because an adapter contains significantly fewer trainable parameters than an entire MLLM. The third approach has the lowest computational cost and is
therefore broadly accessible. To the best of our knowledge, only five PaP methods have been reported within the past year. All of these methods have been evaluated on one or more video question-answering benchmarks and have demonstrated improvements in long-video understanding. However, the methods were evaluated on different benchmarks using different MLLMs. We present a comprehensive evaluation of these five methods using three MLLMs across three long-video understanding benchmarks. Our results show that QAaF achieves the best performance in 13 of the 15 aggregate evaluation settings, while FOCUS ranks second overall. These results provide a common experimental reference for comparing training-free keyframe-selection methods for MLLMs.
\end{abstract}
\keywords{Multimodal model, vision language model, long video understanding, MLLM-agnostic, plug-and-play software, video understanding benchmarks.}

\section{Introduction}
\label{sec:intro}
\begin{figure}[htb]
     \centering
     \includegraphics[width=0.95\linewidth]{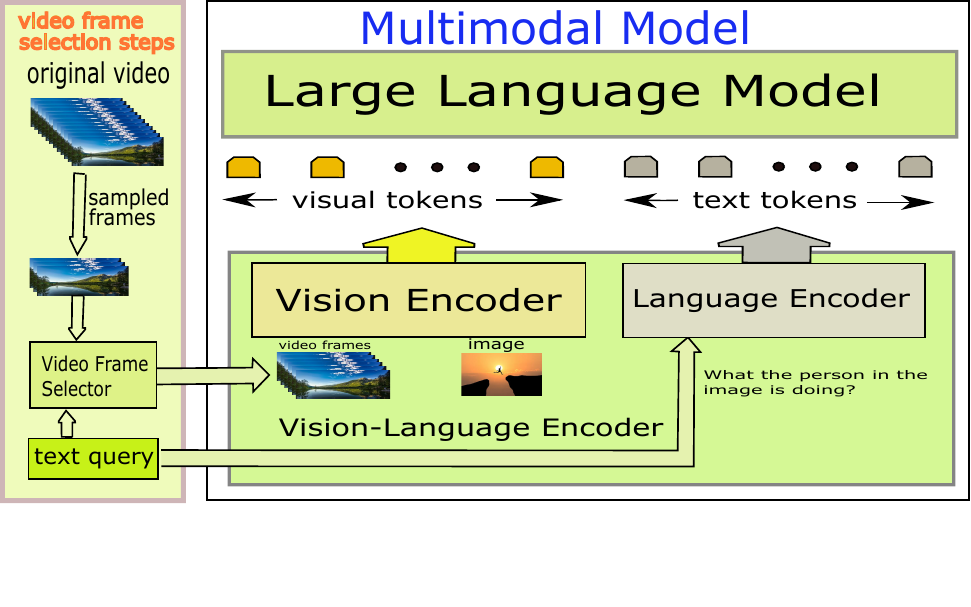}
     \caption{
     The diagram of a video query answering system.
The block on the right represents a multimodal large language model (MLLM), whereas the block on the left represents a video-frame-selection module. The MLLM consists of a front-end module that generates tokens from text, images, and videos, and a back-end module that processes these tokens to generate an answer. The frame-selection module is training-free, MLLM-agnostic, and plug-and-play.}
\label{fig:MLLMandFrameSelector}
 \end{figure}

Multimodal Large Language Models (MLLMs), also known as multimodal models, process both visual and textual inputs for creating textual outputs. As shown in the block on the right of Fig.~\ref{fig:MLLMandFrameSelector}, an MLLM has a front-end module that converts text, image, and video inputs into a sequence of tokens that are then fed into an LLM that generates textual outputs based on these tokens~\cite{lin2023_video_llava,Qwen2VLwang2024qwen2_vl,LLaVAMiniZhang2025llava}. The LLM can process only a limited number of tokens. For example, VideoLLaMA2~\cite{VideoLLaMA2Cheng2024videollama2} can handle around 2,000 tokens, whereas VILA-V1.5~\cite{lin2024vila} supports approximately 4,000 tokens.
 \par
\begin{figure}[htb]
    \centering
    \includegraphics[width=0.95\linewidth]{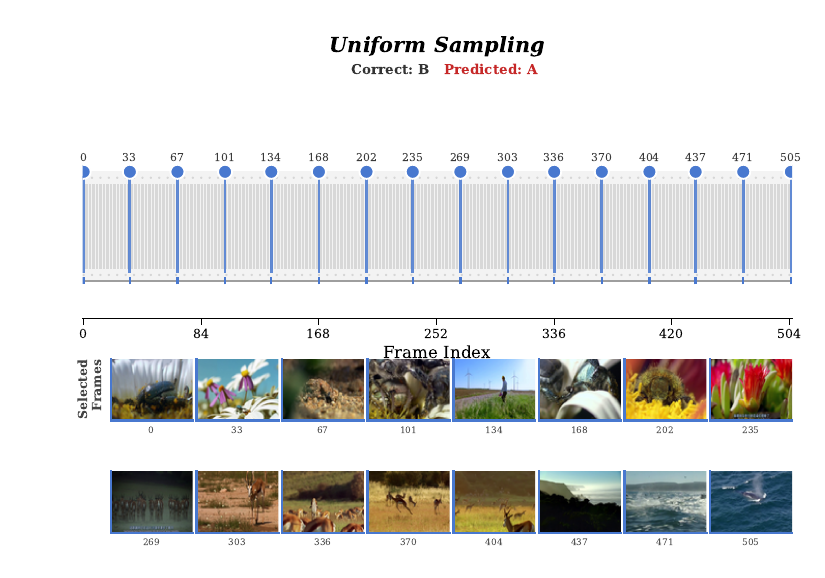}
    \caption{Uniform interval-based frame selection. The 16 frames shown are equally spaced throughout the video.}
    \label{fig:uniformSampling}
\end{figure}
MLLMs have demonstrated excellent image understanding capabilities because the number of visual tokens produced from an image is typically within the processing capacity of the back-end LLM. However, this limited token budget poses a significant challenge for long-video understanding. For example, a 5-minute video recorded at 24 frames per second contains 7,200 frames, which generate far more visual tokens than a back-end LLM can process. To address this problem, the most straightforward approach is to uniformly sample frames at regular intervals, irrespective of the query. However, this approach is prone to excluding frames that contain the critical visual cues needed to answer the query correctly~\cite{tang2025adaptive,QframeZhang2025q,zhu2026focus}. For example, although the 16 uniformly sampled frames shown in Fig.~\ref{fig:uniformSampling} contain diverse visual content, they do not necessarily include the frames most relevant to the query~\cite{Islam2026QueryAlignedVFS}.

\paragraph*{Approaches to Enhance Video Understanding}
\label{sec:relatedWork} 
Three approaches have been proposed to enhance the video-understanding capabilities of MLLMs: 
\begin{inparaenum}[$(i)$] \item \textit{Retraining an MLLM} on a large video corpus~\cite{chen2024internvl,lin2023_sphinx,video_chatgptMaaz2024,wang2022_internvideo} and/or extending its input length; \item \textit{Training an adapter} for a given MLLM that takes the entire video and the query as input and selects the most relevant video frames~\cite{hu2025mllm,wang2023vaquita,LLaVAMiniZhang2025llava,liang2024keyvideollm}; and \item \textit{Developing a training-free plug-and-play (PaP) adapter} that is MLLM-agnostic. The approach is MLLM-agnostic because from a long video the module selects frames  that are most aligned to the query for answering the query and then feeds these keyframes to the MLLM. For the remainder of the paper, we refer to it as the PaP method for keyframe selection. Some PaP methods may select frames solely from the candidate video frames without considering the query~\cite{li2026maxinfo}, but, in general, they use both the candidate frames and the query during frame selection~\cite{QframeZhang2025q,tang2025adaptive,zhu2026focus,Islam2026QueryAlignedVFS}.\end{inparaenum}
\par
The first approach is prohibitively expensive and therefore beyond the reach of most researchers. The second approach, although computationally intensive and time-consuming, is accessible to many research groups because an adapter contains significantly fewer trainable parameters than an entire MLLM. The third approach has the lowest computational cost and is therefore broadly accessible, including to researchers in academic environments. To the best of our knowledge, only five PaP methods have been reported, all within the past year. Interestingly, PaP methods have improved the performance of retrained MLLMs in the first category and are expected to provide similar benefits to future models. Therefore, \textit{the development of PaP methods for query-aligned frame selection is of considerable importance}.

\par
In this paper, we review the five recently proposed PaP methods for long-video understanding: Q-Frame (ICCV 2025)~\cite{QframeZhang2025q}, AKS (Adaptive Keyframe Sampling) (CVPR 2025)~\cite{tang2025adaptive}, FOCUS (ICLR 2026)~\cite{zhu2026focus}, MaxInfo (WACV 2026)~\cite{li2026maxinfo}, and Query-Answer-Aligned Frame Selection (QAaF) (Technical Report, 2026)~\cite{Islam2026QueryAlignedVFS}. Each method has been evaluated on one or more video question-answering benchmarks and has demonstrated improvements in long-video understanding. However, these methods were evaluated on different benchmarks, with different MLLMs, and using different ablation protocols. Table~\ref{tab:video_methods_extended_I} in Sec.~\ref{sec:PaPMethods} summarizes the VLMs used for token generation to feed into MLLMs, the performance of these methods on MLLMs, and the benchmarks used for evaluation. It is easy to see that there are no head-to-head comparisons among the methods. This paper briefly describes all five methods and compares their performance on three advanced MLLMs across three benchmarks.

\par
In this work, we present a comprehensive evaluation of five existing PaP methods across three benchmarks and three MLLMs, and compare their strengths and weaknesses. The remainder of this paper is organized as follows. Section~\ref{sec:PrelimAndProbel} formulates the problem after introducing the necessary notation and definitions. Section~\ref{sec:PaPMethods} describes the PaP methods, and Sec.~\ref{sec:evaluation} presents the empirical evaluation results. Finally, Sec.~\ref{sec:conclusion} provides concluding remarks and discusses potential directions for future research.
\section{Preliminaries and Problem Statement}
\label{sec:PrelimAndProbel} 
\subsection{Preliminaries}
Let $[n]$ be a set of $n$ integer elements.
Throughout this paper, scalars are denoted by non-bold italic letters, vectors by bold lowercase letters,
and matrices by bold uppercase letters.
Let $\mathcal{V}^{(G)} = (\mathbf{V}^{(G)}_i|i \in [T])$ be a video having a sequence of $T$ uncompressed video frames (i.e., images). For $[T_s] \subseteq [T]$, let $\mathcal{V}^{(S)} = (\mathbf{V}^{(G)}_i| i \in [T_s])$ be a subsequence of $T_s$ frames sampled from $\mathcal{V^{(G)}}$. 
Also, for $[k] \subseteq [T_s]$, let 
$\mathcal{V}^{(M)} = (\mathbf{V}^{(S)}_i~|~i~\in~[k])$ 
be a subsequence of $k$ keyframes selected from $\mathcal{V^{(S)}}$.
\par
Let $Q$ denote a text query, possibly associated with $n_a$ answer choices, $A = \{a_i \mid i \in [n_a]\}$. Let each answer choice $a_i \in A$ be paired with $Q$ to form a set of $n_a$ text sequences, $QA = \{Qa_i \mid i \in [n_a]\}$. Let $\mathcal{M}_{{\theta}_{m}}(\cdot,\cdot)$ denote an MLLM that processes $Q$ (or $Qa_i$) together with $\mathcal{V}^{(M)}$ and produces an output $\hat{A}$, i.e., $ \hat{A} = \mathcal{M}_{{\theta}_{m}}(Q,\mathcal{V}^{(M)}).$ Let $\mathcal{VL}_{{\theta}_{vl}}(\cdot)$ denote a pretrained vision-language model, such as CLIP~\cite{radford2021clip}, that converts an input video frame $\mathbf{V}$ into a token vector $\mathbf{vt}$ and a text query $Q$ into a token vector $\mathbf{qt}$. Note that $\mathbf{vt}$ and $\mathbf{qt}$ have the same dimensionality.

\subsection{Problem Statement}
\label{sec:problemStmt}
Given a video $\mathcal{V}^{(G)}$, a text query $Q$, and an MLLM $\mathcal{M}_{\theta_m}(\cdot,\cdot)$ that can process the query $Q$ and at most $k$ video frames from $\mathcal{V}^{(G)}$ to generate an answer $\hat{A}$, the objective is to select a possibly non-consecutive sequence of $k$ keyframes, denoted by $\mathcal{V}^{(M)}$, from $\mathcal{V}^{(G)}$ such that the probability of generating the correct answer to $Q$ is maximized.
\paragraph{Best Frame Subset}
To select the best frame subset, a function for assigning a confidence score to an answer is required. 
Let $R_{\mathcal{M}_{\theta_{m}}}(Q,\mathcal{V}^{(G)})$   be a function that estimates the confidence score produced by the MLLM. Using this function, define an ideal frame selection function $FS^{(ideal)}(Q,\mathcal{V}^{(G)})$  that selects the best frame subset/sequence $\mathcal{V}^{(M)}$ from a given video frame sequence $\mathcal{V}^{(G)}$. A formal definition is given below:
\begin{equation}
FS^{(ideal)}(Q,\mathcal{V}^{(G)}) = \underset{[k] \subseteq [T]} {\arg \max} R_{\mathcal{M}_{{\theta}_{m}}}(Q,\mathcal{V}^{(M)})
\label{eq:bestKeyrameSecetion}
\end{equation}
There are two fundamental challenges associated with the optimal frame-selection function in Eq.~\ref{eq:bestKeyrameSecetion}: \begin{inparaenum}[$(i)$] \item the combinatorial search space is intractable; and \item the confidence-estimation function $R_{\mathcal{M}_{\theta_m}}(\cdot,\cdot)$ is unavailable. \end{inparaenum}
\par
While there is no function $R_{\mathcal{M}_{\theta_{m}}}(Q,\mathcal{V}^{(G)})$ for obtaining an ideal estimate, all PaP methods assume that there exists a function $R_{\mathcal{M}_{\theta_{m}}}^{(1)}(Q,\mathbf{V})$ that can give a good estimate for a single frame $\mathbf{V}$, or can be defined from $\mathbf{vt}$ and $\mathbf{qt}$. For example, AKS and FOCUS use VLMs as $R_{\mathcal{M}_{\theta_{m}}}^{(1)}(Q,\mathbf{V})$, while the other methods define their own scoring functions for this purpose.
\paragraph{Polynomial-time Keyframe Selection}
Using the function $R_{\mathcal{M}_{\theta_{m}}}^{(1)}(Q,\mathbf{V})$, one can evaluate the contributions of all frames in $\mathcal{V}^{(S)}$ for a query $Q$. The AKS and FOCUS methods use VLMs such as CLIP or LongCLIP to obtain an approximation of $R_{\mathcal{M}_{\theta_{m}}}^{(1)}(Q,\mathbf{V})$. The Q-Frame method uses the inner product, whereas  the QAaF method uses cosine similarity between $\mathbf{vt}$ and $\mathbf{qt}$, which are generated by VLMs. The MaxInfo method uses only $\mathbf{vt}$ generated by VLMs. A detailed description of these methods is provided in the next section.

\par
Furthermore, to avoid combinatorial intractability, all PaP methods implicitly assume a frame-level \textit{cooperative} property among video frames, analogous to the contribution of players in an $n$-person cooperative game~\cite{shapley1953value}.
Next, we explicitly state this assumption.

\begin{assumption}
Let $\mathbf{V}_1, \mathbf{V}_2$, and $\mathbf{V}_3$ be three video frames such that 
\[ R_{\mathcal{M}_{\theta_m}}^{(1)}(Q,\mathbf{V}_1) > R_{\mathcal{M}_{\theta_m}}^{(1)}(Q,\mathbf{V}_2) > R_{\mathcal{M}_{\theta_m}}^{(1)}(Q,\mathbf{V}_3). \] It is assumed that \[ R_{\mathcal{M}_{\theta_m}}(Q,(\mathbf{V}_1,\mathbf{V}_2)) > R_{\mathcal{M}_{\theta_m}}(Q,(\mathbf{V}_1,\mathbf{V}_3)). \]
\end{assumption}
The above assumption about three video frames naturally extends to any number of video frames. It says that video frames ordered by
the function $R_{\mathcal{M}_{\theta_{m}}}^{(1)}(Q,\mathbf{V})$ and the above cooperative property allow one to evaluate the contribution of each frame individually. Let $\mathcal{V}^{(S)}_Q = (\mathbf{V}^{(Q)}_i| i \in [T_s])$ be the list of frames in $\mathcal{V}^{(S)}$ sorted in decreasing order of the estimated contributions of the frames for the query $Q$, i.e., $\mathbf{V}^{(Q)}_i \geq \mathbf{V}^{(Q)}_j$ for $i < j$. One can obtain $\mathcal{V}^{(S)}_Q$ after computing alignment scores for all frames in $\mathcal{V}^{(S)}$ and then sorting them in $O(T_s \log T_s)$ time. Now, the first $k$ frames in $\mathcal{V}^{(S)}_Q$ are the desired keyframes. The lemma below describes the selection method.

\begin{lemma}
Given a function $R_{\mathcal{M}_{\theta_{m}}}^{(1)}(Q,\mathbf{V}), \mathcal{V}^{(S)},$ and $Q$,
\begin{equation}
FS^{(est)}(Q,\mathcal{V}^{(S)}) = 
\underset{[k] \subseteq [T]} {\arg \max} R_{\mathcal{M}_{\theta_{m}}}(Q,\mathbf{V}^{(S)}_i) = \sum_{i=1}^k \mathbf{V}^{(Q)}_i
\label{eq:NearBestKeyframeSelection}
\end{equation}
\end{lemma}
\subsection{Video Frame Sampling}
Because the answers to many video-related queries are contained in only a few, often consecutive, video frames~\cite{zhu2026focus}, the query-answering problem can be divided into two subproblems: \begin{inparaenum}[$(i)$] \item selecting the video frames that are relevant to the query and \item generating an answer with an MLLM using the selected keyframes and the query. \end{inparaenum}

 \par
\begin{figure}[htb]
    \centering
    \includegraphics[width=0.95\linewidth]{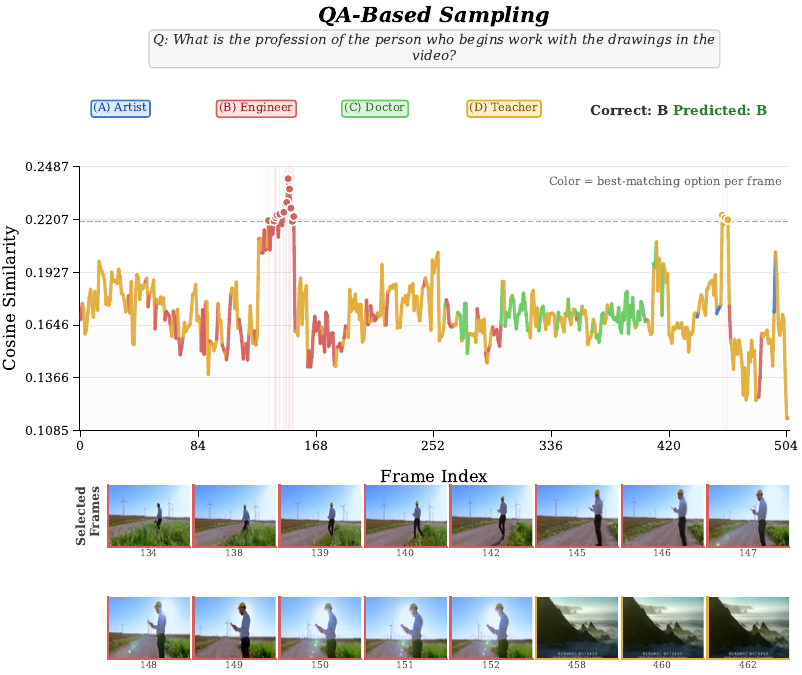}
    \caption{Query-frame-aligned video frame selection. The 16 frames that are most aligned with the query are shown.}
    \label{fig:queryGuidedSelection}
\end{figure}

However, selecting relevant keyframes from a long video can be computationally challenging. Decoding and evaluating every frame requires significant storage space (up to 20 TB for a 2-hour video~\cite{uncompressedWideo2024,poynton2012BookVideoCompression}), CPU, and GPU memory, resulting in longer inference times, particularly for longer videos~\cite{zhu2026focus}. Thus, a practical frame selection method first reduces the candidate frame pool to $T_s$ via fixed-rate sampling~\cite{QframeZhang2025q,tang2025adaptive,zhu2026focus,li2026maxinfo,Islam2026QueryAlignedVFS}; then keyframes are selected from the candidate pool. Fig.~\ref{fig:queryGuidedSelection} shows 16 keyframes with the highest query-aligned scores that were selected using QAaF~\cite{Islam2026QueryAlignedVFS}.

\section{MLLM-Agnostic Training-Free PaP Methods}
\label{sec:PaPMethods}
\par
The block on the left in Fig.~\ref{fig:MLLMandFrameSelector} shows 
a high-level block diagram of a query-aware, training-free, and MLLM-agnostic module of PaP methods 
that is used in Q-Frame~\cite{QframeZhang2025q}, AKS~\cite{tang2025adaptive}, FOCUS~\cite{zhu2026focus}, and QAaF~\cite{Islam2026QueryAlignedVFS} for selecting keyframes. There are two independent yet synergistic modules: one for video frame subsampling to reduce the candidate frame pool, and the other for query-aware video frame selection from the reduced pool.
\par
For frame selection, AKS and FOCUS use a VLM to compute frame-query alignment scores and then select the $k$ frames with the highest scores. In contrast, Q-Frame and QAaF first use a VLM to obtain frame and text embeddings and then compute alignment scores using their respective algorithms. MaxInfo~\cite{li2026maxinfo} is query-agnostic: it uses the embeddings of all frames in the candidate pool and selects the frames that are most mutually dissimilar.
\par
Before we present a concise description of the PaP methods, essential definitions are presented next.

\begin{definition}[Inner product]
    \label{def:innerProd}
  Let $\mathbf{x}$ and $\mathbf{y}$ be two nonzero vectors of identical length. Their inner product is defined as $ \operatorname{InnProd}(\mathbf{x},\mathbf{y}) = (\mathbf{x}^{\top}\mathbf{y})$.
\end{definition}
The inner product measures the projection length of the vector $\mathbf{y}$ in the direction of the vector $\mathbf{x}$.
\begin{definition}[Cosine similarity]
    \label{def:cosSim}
Let $\mathbf{x}$ and $\mathbf{y}$ be two nonzero vectors of identical length. Their cosine similarity is defined as $ \operatorname{CosSim}(\mathbf{x},\mathbf{y}) = (\mathbf{x}^{\top}\mathbf{y})/(\lVert\mathbf{x}\rVert_2 \lVert\mathbf{y}\rVert_2)$, where both vectors are normalized using the $\ell_2$ norm.
\end{definition}
The $\operatorname{CosSim}$ measures the cosine of the angle between the vectors $\mathbf{y}$ and $\mathbf{x}$; the smaller the angle, the larger the cosine value.

\begin{table*}[htb]
\caption{Comparison of video frame selection methods for video understanding systems.
}
\centering
\small
\renewcommand{\arraystretch}{1.3}
\begin{tabular}{|p{1.0cm}|p{2.0cm}|p{2.95cm}|p{3.2cm}|p{5.0cm}|}
\hline
\textbf{Method} & \textbf{VLMs} & \textbf{MLLMs} & \textbf{Benchmarks} & \textbf{Ablation Studies} \\
\hline

Q-Frame&
CLIP &
VILA-V1.5, GPT-4, Qwen2-VL &
MLVU, Video-MME, LongVideoBench &
Frame sampling rate, resolution effects, multi-resolution adaptation, resolution distribution \\

\hline
AKS
&
CLIP, BLIP,  SeViLA &
GPT-4, Qwen2-VL, LLaVA-Video &
Video-MME, LongVideoBench  &
Frame sampling rate, influence of different VLMs, hyperparameter sensitivity \\

\hline
FOCUS
&
CLIP, BLIP, SigLIP &
GPT-4, Qwen2-VL, LLaVA-Video &
Video-MME, VSI-Bench, LongVideoBench &
Two-stage exploration strategy, Bernstein confidence radius, clip length variations\\

\hline

MaxInfo&
CLIP, SigLIP, DINOv2 &
LLaVA-Video, Qwen2-VL, InternVL2 &
Video-MME &
Influence of VLM choice, effect of hyperparameters \\

\hline

QAaF&
CLIP, LongCLIP &
LLaVA-Video, Qwen2-VL
&
Video-MME, MLVU, LongVideoBench &
Number of frames selected for MLLM input, effect of frame budget \\

\hline

\end{tabular}
\label{tab:video_methods_extended_I}
\end{table*}

\subsection{Known Keyframe Selection Methods}
\label{sec:SolutionMethods}
\paragraph{Q-Frame Method~\cite{QframeZhang2025q}}
The Q-Frame method uses video and query tokens. For each video frame $\mathbf{V}_i^{(S)}\in\mathcal{V}^{(S)}$, a VLM generates
a frame token vector $\mathbf{vt}_i$. The VLM also generates a query token vector $\mathbf{qt}$ for the query $Q$. Q-Frame
computes $|T_s|$ inner products, $\operatorname{InnProd}(\mathbf{qt},\mathbf{vt}_i)$, from the $T_s$ video tokens and then uses the Gumbel trick~\cite{jang2016gumbel_softmax} to select the $k$ most aligned frames. For MLLMs that support multi-resolution inputs, such as Qwen2-VL, Q-Frame divides the selected frames into high-, regular-, and low-resolution groups and feeds them, together with the query, into the MLLM.

\par 
To evaluate Q-Frame, the candidate frame pool was created by sampling original video frames at regular intervals. The paper used CLIP as the VLM. The performance of the method was evaluated on three MLLMs (VILA-V1.5, GPT-4, and Qwen2-VL) using the MLVU, LongVideoBench (12 min), and Video-MME benchmarks. For Video-MME, average (17 min), short (1.3 min), medium (9 min), and long (41 min) video categories are reported.
\par
For determining the contribution of different steps, the ablation study included the frame sampling rate, fixed- and multi-resolution frames, and different distributions of high-, regular-, and low-resolution frames.

\paragraph{FOCUS  Method~\cite{zhu2026focus}} 
The FOCUS (Frame-Optimistic Upper-Bound Selection) method utilizes an empirical observation that the temporal autocorrelation of the per-frame relevance of a query spans several seconds for two video understanding benchmarks. Based on this observation, the method models a video as multiple arms of a multi-armed bandit. Each video segment is considered an arm of the bandit, and the relevance of each arm for containing keyframes is initially estimated by sampling a few frames in the segment. Segments containing frames with higher relevance scores are then evaluated further to select keyframes.
\par
To evaluate FOCUS, the entire video is divided into equal-length segments. Sampled frames from each video segment are evaluated using a VLM that takes a frame and the query as inputs and produces a relevance score. For the empirical evaluation of the method, frame-query relevance was computed using the BLIP, SigLIP, and CLIP VLMs. The performance of the method was evaluated on three MLLMs (GPT-4, Qwen2-VL, and LLaVA-Video) using the Video-MME (short, medium, and long), LongVideoBench (short, medium, and long), and VSI-Bench benchmarks.
\par
The ablation study examined the effects of two-stage exploration, the Bernstein confidence radius, and clip lengths
of 8, 16, and 32 seconds.

\paragraph{ Adaptive Keyframe Sampling (AKS) Method~\cite{tang2025adaptive}}
As shown below, the AKS method adds a term to Eq.~\ref{eq:NearBestKeyframeSelection} to extend the temporal range of the selected frames.
\begin{equation}
\begin{medsize}
FS^{(est)}(Q,\mathcal{V}^{(S)}) = \underset{[k] \subseteq [T]} {\arg \max} \sum_{i\in [k]} R_{\mathcal{M}_{\theta_{m}}}(Q,\mathbf{V}^{(G)}_i) + \lambda c([k])
\label{eq:AksBestKeyframeSelection}
\end{medsize}
\end{equation}
When $\lambda = 0$, frames with the highest relevance are selected. The additional term generalizes the objective by widening the temporal range of the selected frames and may improve performance when answering the query requires information from different video segments. The AKS paper computes relevance scores using a VLM that provides a rating score for a given input video and query. For experimental evaluation, three VLMs were used (BLIP, CLIP, and SeViLA). AKS was evaluated with GPT-4, Qwen2-VL, and LLaVA-Video on the LongVideoBench and Video-MME benchmarks.

\par
The ablation study examined the sampling rate, the choice of VLM, and the model hyperparameters.
\paragraph{MaxInfo Method}
MaxInfo attempts to select the $r$ most mutually dissimilar keyframes from a given video (irrespective of the query to be answered). It does so in three sequential steps. First, $n$ candidate frames are selected by sampling frames at regular intervals. In the second step, using a VLM, each candidate frame is converted to a $d$-dimensional feature vector. These feature vectors form an $n \times d$ matrix $\mathbf{Q}$, which is converted to an $n \times s$ matrix $\mathbf{Q}_s$ using truncated singular value decomposition, where $s < d$. Because $\mathbf{Q}_s$ is created using the $s$ largest singular values, the most significant visual variations are retained. In the last step, $r$ rows of $\mathbf{Q}_s$ are selected using the rectangular MaxVol algorithm from~\cite{mikhalev2018rectangulaMaxVol}. The video frames corresponding to these $r$ rows are assumed to contain the most information in the video. It is important to note that MaxInfo's frame selection is query-agnostic.

\par
For empirical evaluation, three VLMs (CLIP, DINOv2, and SigLIP) were used to generate frame token vectors. MaxInfo was evaluated with three MLLMs (LLaVA-Video, Qwen2-VL, and InternVL2) on the Video-MME benchmark. The effects of the VLMs and the hyperparameters of the MaxVol algorithm were studied.

\paragraph{Query-Answer-Aligned Frame Selection (QAaF)} 
The QAaF method uses cosine similarity to measure the alignment of (1) image and query vectors produced by VLMs, and (2) image and concatenated query-answer vectors produced by VLMs. The authors believe that cosine similarity better measures the alignment between the query token vector $\mathbf{qt}$ and the frame token vector $\mathbf{vt}$. 

There are two versions of the algorithm. The first version considers only the query for alignment, similar to the Q-Frame, FOCUS, and AKS methods. The second version is designed for multiple-choice questions. Each answer is appended to the query to obtain $n_a$ query-answer texts. The alignment of each of these texts with $n$ frame vectors is computed, generating $n \cdot n_a$ alignment scores. From these scores, the $k$ frames with the highest alignment scores are selected. Both variants were evaluated using CLIP and LongCLIP with Qwen2-VL, LLaVA-Video, and InternVL3.5 on the Video-MME, LongVideoBench, and MLVU benchmarks.

\section{Experimental Evaluation}
\label{sec:evaluation}

\subsection{Experimental Setup}
\label{sec:expSetup}
We evaluated all five keyframe selection methods using three public benchmarks ({MLVU}, {Video-MME}, and {LongVideoBench}) on three MLLMs ({Qwen2-VL-7B}~\cite{Qwen2VLwang2024qwen2_vl}, {LLaVA-Video-7B}~\cite{zhang2024llava-video}, and {InternVL3.5}~\cite{chen2024internvl}). These \textbf{benchmarks cover} different video \textbf{durations, domains, and reasoning skills}. {MLVU} consists of long videos covering seven closed-ended categories: Topic Reasoning (TR), Anomaly Recognition (AR), Needle QA (NQA), Ego Reasoning (ER), Plot QA (PQA), Action Order (AO), and Action Count (AC). {Video-MME} assesses video understanding across diverse domains and video lengths: short, medium, and long. {LongVideoBench} focuses on understanding long-context videos of various durations, with questions that require reasoning across multiple video segments. 

\textbf{Computing Resources:} Experiments were conducted on a Linux server equipped with four NVIDIA H100 NVL GPUs, each with 96 GB of GPU memory, dual Intel Xeon Silver 4514Y CPUs, and 1.5 TB of system memory. 

{\textbf{Fair Comparison Protocol:}} 
For a fair comparison, we base our evaluations on the released evaluation code of all PaP methods whenever possible, making only the necessary adaptations to run the same benchmarks, model backbones, and local data paths. Specifically, we \begin{inparaenum}[$(i)$] \item maintain a fixed budget of $k=16$ selected frames for the uniform sampling, AKS, FOCUS, MaxInfo, and QAaF methods, \item maintain the multi-resolution setting ($4+8+32$ frames) for Q-Frame, \item use BLIP-based frame selection for AKS and FOCUS, and CLIP-family encoders (CLIP or LongCLIP) for Q-Frame, MaxInfo, and QAaF, \item report the same seven closed-ended MLVU task categories for all MLVU comparisons, and \item provide all duration splits, both with and without subtitles, for Video-MME and LongVideoBench. \end{inparaenum}

\begin{table}[htb]
\centering
\renewcommand{\arraystretch}{1.7}
\caption{Summary of the five PaP methods across three MLLMs and three benchmarks. The best result is shown in bold, and the
second-best result is underlined.}
\label{tab:Summary5PaPs3MLLs3Becnsh}
\small
\setlength{\tabcolsep}{2pt}
\begin{tabular}{|p{.3cm}
|p{0.135\columnwidth}|p{0.135\columnwidth}|p{0.135\columnwidth}|p{0.135\columnwidth}|p{0.135\columnwidth}|p{0.135\columnwidth}|}
\hline &
\textbf{Datasets} &
\textbf{Q-Frame} &
\textbf{AKS} &
\textbf{FOCUS} &
\textbf{MaxInfo} &
\textbf{QAaF} \\
\hline
\multirow{3}{*}{\rotatebox[origin=c]{90}{Qwen2-VL}}
&MLVU &56.3& 62.7&\underline{66.19}& 62.2 &\textbf{69.2} \\
\cline{2-7}
&Video-MME& 56.96\, /\,\underline{60.89} & 55.19\,/\,54.70 & \underline{57.07}\,/\,59.33 & 55.96\,/\,55.78 & \textbf{60.90}\,/\,\textbf{62.93}\\
\cline{2-7}
&Long VBench & 57.29\,/\,57.44 & 54.90\,/\,54.45 & \underline{58.04}\,/\,\underline{58.19} & 53.48\,/\,54.53 & \textbf{58.80}\,/\,\textbf{60.20} \\
\hline \cline{1-7}
\multirow{3}{*}{\rotatebox[origin=c]{90}{LLaVA-Video}}
&MLVU & \underline{70.4} & 65.5 & 69.0 & 67.2 & \textbf{71.3} \\
\cline{2-7}
&Video-MME& 60.44\,/\,\underline{63.89} & \textbf{61.93}\,/\,63.11 & 60.52\,/\,60.70 & 60.41\,/\,60.37 & \underline{61.81}\,/\,\textbf{64.70}\\
\cline{2-7}
&Long VBench & 59.98\,/\,\underline{60.51} & \underline{60.21}\,/\,59.69 & 58.86\,/\,60.06 & 57.52\,/\,57.59 & \textbf{61.32}\,/\,\textbf{63.49} \\
\hline \cline{1-7}
\multirow{3}{*}{\rotatebox[origin=c]{90}{InternVL3.5 }}
&MLVU & 68.29 & 66.82 & \underline{69.12} & 67.7 & \textbf{70.18} \\
\cline{2-7}
&Video-MME& 60.33\,/\,63.30 & 58.59\,/\,61.52 & \textbf{61.56}\,/\,\underline{63.37} & 59.33\,/\,62.00 & \underline{60.89}\,/\,\textbf{64.52}\\
\cline{2-7}
&Long VBench & \underline{60.58}\,/\,61.71 & 59.76\,/\,62.53 & 60.28\,/\,\underline{62.68} & 58.41\,/\,61.03 & \textbf{63.20}\,/\,\textbf{66.57} \\
\hline
\end{tabular}
\end{table} 

\subsection{Main Results}
\label{sec:MainResults}

Tables~\ref{tab:mlvu_results}, \ref{tab:videoMME_results}, and \ref{tab:LongVideoBench_results} reports  results for  all PaP methods: Q-frame~\cite{QframeZhang2025q}, AKS~\cite{tang2025adaptive}, FOCUS~\cite{zhu2026focus}, MaxInfo~\cite{li2026maxinfo}, QaF and QAaF~\cite{Islam2026QueryAlignedVFS} and the uniform-sampling baseline across three MLLMs and three benchmarks. 
The best result in each comparison is shown in bold. Because these complete tables have a total of 546 entries, it is quite challenging to compare them, Table~\ref{tab:Summary5PaPs3MLLs3Becnsh} provides a concise summary of the five PaP methods. For Video-MME and LongVideoBench, abbreviated as Long VBench in Table~\ref{tab:Summary5PaPs3MLLs3Becnsh}, each cell reports
results without and with subtitles, respectively.

Table~\ref{tab:Summary5PaPs3MLLs3Becnsh}
 shows that, overall,the QAaF method performs considerably better than the other methods. Across the 15 aggregate evaluation settings, it achieves the best performance in 13 cases and the second-best performance in the remaining two. The FOCUS method ranks second overall, with one best performance and seven second-best performances. Among the MLLMs, InternVL3.5 achieves the best performance on Video-MME and LongVideoBench, whereas LLaVA-Video achieves the best performance on MLVU.

\subsubsection{Results for MLVU benchmark dataset }
\label{mlvuResults}

\begin{table}[htb]
\caption{Comparison of different methods on MLVU. 
The best performance is boldfaced, and the second-best performance is underlined.}
\small
\setlength{\tabcolsep}{5pt}
\renewcommand{\arraystretch}{1.08}
\resizebox{\textwidth}{!}{%
\begin{tabular}{llccccccccc}
\toprule
Methods & Size & \#Frames & TR & AR & NQA & ER & PQA & AO & AC & Avg. \\
\specialrule{0.08em}{0em}{0em}
Qwen2-VL (Uniform) & 7B & 16 & \textbf{86.7} & \textbf{71.5} & 72.4 & 56.3 & 60.9 & 42.9 & 22.8 & 60.4 \\
Qwen2-VL + Q-Frame & 7B & 4+8+32 & 76.8 & 57.0 & 74.6 & 48.0 & 59.9 & 40.5 & 22.3 & 56.3 \\
Qwen2-VL + FOCUS  & 7B & 16 & 80.99 & 65.00 & 81.13 & 62.78 & 69.76 & 47.88 & \textbf{41.75} & 66.19 \\
Qwen2-VL + AKS & 7B & 16 & 82.1 & 64.0 & 78.9 & 56.8 & 63.3 & 44.4 & \uline{40.3} & 62.7 \\
Qwen2-VL + MaxInfo & 7B & 16 & 83.3 & 69.5 & 78.0 & 58.8 & 62.5 & 42.1 & 31.1 & 62.2 \\

Qwen2-VL + QaF + LongCLIP & 7B & 16
& \uline{84.4}
& 67.5
& \textbf{87.0}
& \uline{65.1}
& \uline{71.6}
& \uline{50.6}
& 28.6
& \uline{67.7} \\
Qwen2-VL + QAaF + LongCLIP & 7B & 16 & 81.0 & \uline{70.0} & \uline{86.2} & \textbf{67.6} & \textbf{75.0} & \textbf{55.2} & 29.6 & \textbf{69.2} \\

\midrule
LLaVA-Video + Uniform & 7B & 16
& 81.0
& \uline{59.5}
& 77.2
& 61.4
& 68.8
& 44.8
& 34.5
& 63.5 \\

LLaVA-Video + Q-Frame & 7B & 4+8+32
& \uline{82.9}
& 52.0
& 81.4
& 68.5
& \uline{77.4}
& \textbf{59.8}
& \textbf{51.9}
& \uline{70.4} \\

LLaVA-Video + FOCUS & 7B & 16
& 81.0
& 59.0
& \uline{83.4}
& 69.0
& 73.8
& 49.8
& \uline{49.5}
& 69.0 \\

LLaVA-Video + AKS & 7B & 16
& 82.1
& 57.5
& 77.7
& 61.9
& 70.3
& 45.9
& 48.5
& 65.5 \\

LLaVA-Video + MaxInfo & 7B & 16
& \textbf{84.8}
& \textbf{65.5}
& 82.8
& 65.1
& 65.5
& 52.9
& 45.6
& 67.2 \\

LLaVA-Video + QaF + LongCLIP & 7B & 16
& 81.7
& 50.5
& \textbf{84.5}
& \uline{70.7}
& 75.0
& 49.4
& 47.6
& 68.8 \\

LLaVA-Video + QAaF + LongCLIP & 7B & 16
& 79.1
& 57.0
& \textbf{84.5}
& \textbf{71.0}
& \textbf{80.1}
& \uline{57.1}
& 48.1
& \textbf{71.3} \\

\midrule
InternVL3.5 + Uniform & 8B & 16 & 82.13 & 67.00 & 80.28 & 62.93 & 64.56 & 51.35 & 37.38 & 65.07 \\

InternVL3.5 + AKS & 8B & 16 & 80.99 & \uline{71.00} & 81.41 & 59.77 & 67.90 & 51.74 & 47.57 & 66.82 \\

InternVL3.5 + FOCUS & 8B & 16 & 79.47 & 61.50 & \textbf{84.23} & \textbf{67.82} & \textbf{71.99} & 56.76 & 47.57 & \uline{69.12} \\

InternVL3.5 + Q-Frame & 8B & 4+8+32 & \textbf{84.03} & 64.50 & 81.69 & \uline{64.94} & 69.94 & 56.37 & 45.15 & 68.29 \\

InternVL3.5 + MaxInfo & 8B & 16 & \uline{82.5} & 67.5 & \uline{84.2} & 64.4 & 67.2 & 56.0 & 41.8 & 67.7 \\

InternVL3.5 + QaF + LongCLIP & 8B & 16 & 72.91 & \textbf{86.20} & 66.38 & 40.29 & 56.37 & \textbf{60.50} & \textbf{76.43} & 68.25 \\

InternVL3.5 + QAaF + LongCLIP & 8B & 16 & 75.88 & \textbf{86.20} & 66.09 & 39.81 & \uline{70.27} & \uline{57.50} & \uline{75.67} & \textbf{70.18} \\

\bottomrule
\end{tabular}%
}
\label{tab:mlvu_results}
\end{table}
Table~\ref{tab:mlvu_results} compares uniform sampling and six frame-selection methods across three MLLMs on MLVU. QAaF achieves the best average performance for all three models, improving Qwen2-VL, LLaVA-Video, and InternVL3.5 over uniform sampling by 8.8, 7.8, and 5.11 points, respectively. LLaVA-Video with QAaF obtains the highest overall score of 71.3, followed by LLaVA-Video with Q-Frame at 70.4. Across the three MLLMs, QAaF achieves the highest mean performance of 70.23, demonstrating that incorporating both query and answer information into video-frame alignment generalizes across model architectures. Nevertheless, the task-level results show that no single method dominates every category. Uniform sampling remains strongest on TR, QaF obtains the best NQA, AO, and AC results, and QAaF performs particularly well on ER and PQA. The results therefore indicate that answer-aware frame selection improves overall performance but introduces task-dependent trade-offs, especially for InternVL3.5.
\par
We can see that QAaF has the highest cross-model mean, beating QaF by approximately 1.98 points and FOCUS by approximately 2.12 points. QaF is the most stable method numerically, with a standard deviation of only 0.55, but its average performance is lower than QAaF. Q-Frame is highly model-dependent. It performs poorly with Qwen2-VL but very well with LLaVA-Video.

\subsubsection{Results for Video-MME benchmark dataset }
\label{VideoMMEResults}

\begin{table}[htb]
\centering
\caption{Video-MME results (without / with subtitles). The best performance is boldfaced, and the second-best performance is underlined.}
\label{tab:videomme}
\setlength{\tabcolsep}{3pt}
\resizebox{\textwidth}{!}{%
\begin{tabular}{l c c cccc}
\toprule
\multirow{2}{*}{\textbf{Model}} & \textbf{LLM} & \multirow{2}{*}{\textbf{\#Frames}} & \multicolumn{4}{c}{\textbf{Video-MME} (\textit{wo\,/\,w subs})} \\
\cmidrule(lr){4-7}
 & \textbf{Size} & & Overall & Short & Medium & Long \\
 & & & \textit{17min} & \textit{1.3min} & \textit{9min} & \textit{41min} \\
\midrule
\multicolumn{7}{l}{\textit{Fixed input tokens:}} \\
Qwen2-VL + Uniform & 7B & 16 & 56.07\,/\,57.44 & 68.22\,/\,68.30 & 52.33\,/\,55.60 & 47.67\,/\,48.40 \\
Qwen2-VL + Q-Frame  & 7B & 4+8+32 & 56.96\,/\,60.89 & 68.33\,/\,\uline{72.33} & \uline{55.44}\,/\,59.11 & 47.11\,/\,51.22 \\
Qwen2-VL + AKS & 7B & 16 & 55.19\,/\,54.70 & 64.44\,/\,64.11 & 53.22\,/\,53.00 & 47.89\,/\,47.00 \\
Qwen2-VL + FOCUS & 7B & 16 & 57.07/59.33 & 67.56/68.44 & 54.44/56.67 & \uline{49.22}/52.89 \\
Qwen2-VL + MaxInfo & 7B & 16 & 55.96\,/\,55.78 & 66.67\,/\,66.89 & 52.67\,/\,52.33 & 48.56\,/\,48.11 \\
Qwen2-VL + QaF  & 7B & 16 & \uline{57.11}\,/\,\uline{61.04} & \uline{68.78}\,/\,70.22 & 54.33\,/\,\uline{59.89} & 48.22\,/\,\uline{53.00} \\
Qwen2-VL + QAaF + LongCLIP & 7B & 16 & \textbf{60.9}\,/\,\textbf{62.93} & \textbf{70.7}\,/\,\textbf{72.67} & \textbf{60.8}\,/\,\textbf{62.56} & \textbf{51.1}\,/\,\textbf{53.56} \\

\midrule
LLaVA-Video (Uniform) & 7B & 16 & 59.56/62.07 & 71.33/74.11 & 58.22/60.22 & 49.11/51.89 \\
LLaVA-Video +Q-Frame & 7B & 4+8+32 & 60.44\,/\,\uline{63.89} & 72.22\,/\,74.33 & 57.89\,/\,\uline{62.00} & 51.22\,/\,\textbf{55.33} \\
LLaVA-Video +AKS & 7B & 16 & \textbf{61.93}\,/\,63.11 & \uline{73.11}\,/\,\uline{74.67} & \uline{60.89}\,/\,61.00 & \uline{51.78}\,/\,53.67 \\
LLaVA-Video +FOCUS & 7B & 16 & 60.52\,/\,60.70 & 70.40\,/\,69.70 & 58.80\,/\,59.00 & \textbf{52.30}\,/\,53.40 \\
LLaVA-Video + MaxInfo & 7B & 16 & 60.41\,/\,60.37 & 72.67\,/\,72.67 & 57.33\,/\,57.00 & 51.22\,/\,51.44 \\
LLaVA-Video + QaF & 7B & 16 & 58.07\,/\,62.15 & 69.56\,/\,72.78 & 56.33\,/\,60.78 & 48.33\,/\,52.89 \\
LLaVA-Video + QAaF + LongCLIP & 7B & 16 & \uline{61.81}\,/\,\textbf{64.70} & \textbf{73.22}\,/\,\textbf{76.67} & \textbf{61.11}\,/\,\textbf{63.56} & 51.11\,/\,\uline{53.89} \\

\midrule
InternVL3.5 + Uniform & 8B & 16 & 59.41\,/\,62.41 & 69.78\,/\,74.67 & 59.33\,/\,60.11 & 49.11\,/\,52.44 \\

InternVL3.5 + AKS & 8B & 16 & 58.59\,/\,61.52 & 68.33\,/\,70.67 & 57.22\,/\,60.44 & \uline{50.22}\,/\,53.44 \\

InternVL3.5 + FOCUS & 8B & 16 & \textbf{61.56}\,/\,\uline{63.37} & 71.44\,/\,72.78 & \uline{61.22}\,/\,61.11 & \textbf{52.00}\,/\,\textbf{56.22} \\

InternVL3.5 + Q-Frame & 8B & 4+8+32 & 60.33\,/\,63.30 & \uline{72.11}\,/\,\uline{75.44} & 59.56\,/\,\uline{61.78} & 49.33\,/\,52.67 \\

InternVL3.5 + MaxInfo & 8B & 16 & 59.33\,/\,62.00 & 69.78\,/\,72.00 & 59.89\,/\,60.22 & 48.33\,/\,53.78 \\

InternVL3.5 + QaF + LongCLIP & 8B & 16 & 58.85\,/\,61.85 & 71.11\,/\,73.00 & 57.44\,/\,59.00 & 48.00\,/\,53.56 \\

InternVL3.5 + QAaF + LongCLIP & 8B & 16 & \uline{60.89}\,/\,\textbf{64.52} & \textbf{72.89}\,/\,\textbf{75.78} & \textbf{61.67}\,/\,\textbf{63.78} & 48.11\,/\,\uline{54.00} \\

\bottomrule
\end{tabular}%
}
\label{tab:videoMME_results}
\end{table}
From Table~\ref{tab:videoMME_results}, it can be seen that without subtitles, LLaVA-Video + AKS achieves the highest overall score of 61.93, followed by LLaVA-Video + QAaF + LongCLIP (61.81) and InternVL3.5 + FOCUS (61.56). With subtitles, LLaVA-Video + QAaF + LongCLIP achieves the best overall performance of 64.70, narrowly outperforming InternVL3.5 + QAaF + LongCLIP (64.52). QAaF improves the overall score over uniform sampling by 4.83/5.49, 2.25/2.63, and 1.48/2.11 points without/with subtitles for Qwen2-VL, LLaVA-Video, and InternVL3.5, respectively. These results demonstrate that QAaF consistently improves all three MLLMs and is particularly effective when subtitle information is available.
\par
Performance generally decreases as video duration increases, highlighting the difficulty of understanding long videos under a fixed token budget. LLaVA-Video + QAaF achieves the best short-video performance both without and with subtitles (73.22/76.67), while InternVL3.5 + QAaF obtains the best medium-video results (61.67/63.78). For long videos, LLaVA-Video + FOCUS performs best without subtitles (52.30), whereas InternVL3.5 + FOCUS achieves the highest score with subtitles (56.22). Subtitles improve most configurations, with particularly large overall gains for QaF, Q-Frame, and QAaF, although small degradations occur for some AKS, FOCUS, and MaxInfo configurations. Overall, QAaF provides the strongest performance across short and medium videos, while FOCUS is more effective for long-video understanding.

\subsubsection{Results for LongVideoBench benchmark dataset }
\label{LVBenchResults}

\begin{table}[htb]
\centering
\caption{LongVideoBench results (without / with subtitles). The best performance is boldfaced, and the second-best performance is underlined.}
\label{tab:longvideobench}
\setlength{\tabcolsep}{3pt}
\resizebox{\textwidth}{!}{%
\begin{tabular}{l c c ccccc}
\toprule
\multirow{2}{*}{\textbf{Model}} & \textbf{LLM} & \multirow{2}{*}{\textbf{\#Frames}} & \multicolumn{5}{c}{\textbf{LongVideoBench} (\textit{wo\,/\,w subs})} \\
\cmidrule(lr){4-8}
 & \textbf{Size} & & Overall & 8s--15s & 15s--60s & 3m--10m & 15m--60m \\
\midrule
Qwen2-VL (+Uniform) & 7B & 16 & 54.10\,/\,55.80 & \uline{66.70}\,/\,65.10 & 65.10\,/\,68.00 & 52.70\,/\,57.50 & 47.50\,/\,47.70 \\
Qwen2-VL + FOCUS & 7B & 16 & 58.04\,/\,58.19 & 66.14\,/\,\textbf{67.73} & 66.28\,/\,66.86 & 56.07\,/\,58.01 & \textbf{54.26}\,/\,52.48 \\
Qwen2-VL +AKS & 7B & 16 & 54.90\,/\,54.45 & 62.43\,/\,60.85 & 61.63\,/\,62.21 & 56.55\,/\,54.85 & 49.11\,/\,49.65 \\
Qwen2-VL + Q-Frame  & 7B & 4+8+32 & 57.29/57.44 & 62.96\,\,/63.49 & \textbf{73.26}\,/\,69.19 & \uline{58.25}/\uline{59.47} & 49.82/50.35 \\
Qwen2-VL + MaxInfo & 7B & 16 & 53.48\,/\,54.53 & 65.08\,/\,65.61 & 68.02\,/\,69.19 & 52.43\,/\,53.64 & 45.92\,/\,46.99 \\
Qwen2-VL + QaF + LongCLIP & 7B & 16 & \uline{58.12}\,/\,\uline{59.01} & \textbf{68.25}\,/\,\uline{65.61}& 69.19\,/\,\uline{70.93} & 57.52\,/\,\uline{59.47} & \uline{51.77}\,/\,\uline{52.84} \\
Qwen2-VL + QAaF + LongCLIP & 7B & 16 & \textbf{58.80}\,/\,\textbf{60.20} & 65.60\,/\,65.10 & \uline{69.80}\,/\,\textbf{72.7} & \textbf{59.70}\,/\,\textbf{61.20} & \uline{52.50}\,/\,\textbf{54.10} \\

\bottomrule
LLaVA-Video + Uniform & 7B & 16
& 56.90\,/\,59.30
& 67.20\,/\,\uline{69.31}
& 70.35\,/\,72.09
& 53.64\,/\,57.28
& 51.69\,/\,53.48 \\

LLaVA-Video + Q-Frame & 7B & 4+8+32
& 59.98\,/\,60.51
& 67.20\,/\,66.67
& 70.93\,/\,70.35
& 57.28\,/\,58.74
& \uline{56.21}\,/\,56.74 \\

LLaVA-Video + FOCUS & 7B & 16
& 58.86\,/\,60.06
& 62.96\,/\,64.55
& 68.61\,/\,67.44
& 58.25\,/\,59.71
& 54.97\,/\,56.56 \\

LLaVA-Video + MaxInfo & 7B & 16
& 57.52\,/\,57.59
& \uline{67.72}\,/\,\textbf{70.37}
& 69.77\,/\,70.35
& 54.61\,/\,56.31
& 52.48\,/\,50.35 \\

LLaVA-Video + AKS & 7B & 16
& \uline{60.21}\,/\,59.69
& \textbf{70.90}\,/\,67.72
& 69.77\,/\,73.26
& \uline{58.74}\,/\,58.01
& 54.79\,/\,54.08 \\

LLaVA-Video + QaF + LongCLIP & 7B & 16
& 59.99\,/\,\uline{62.83}
& 67.20\,/\,\uline{69.31}
& \textbf{73.26}\,/\,\textbf{75.00}
& 58.01\,/\,\uline{60.92}
& 54.96\,/\,\textbf{58.33} \\

LLaVA-Video + QAaF + LongCLIP & 7B & 16
& \textbf{61.32}\,/\,\textbf{63.49}
& 67.20\,/\,\uline{69.31}
& \uline{72.09}\,/\,\uline{73.84}
& \textbf{59.71}\,/\,\textbf{63.59}
& \textbf{57.22}\,/\,\uline{58.29} \\
\midrule
InternVL3.5 (+Uniform) & 8B & 16 & 58.71\,/\,61.78 & 68.78\,/\,\textbf{73.54} & \textbf{76.16}\,/\,73.84 & 56.55\,/\,60.68 & 51.60\,/\,54.96 \\

InternVL3.5 + AKS & 8B & 16 & 59.76\,/\,62.53 & 63.49\,/\,71.43 & 68.60\,/\,65.70 & 59.47\,/\,63.83 & \uline{56.03}\,/\,\uline{57.62} \\

InternVL3.5 + FOCUS & 8B & 16 & 60.28\,/\,62.68 & 61.38\,/\,66.14 & 68.60\,/\,73.26 & \uline{63.35}\,/\,64.81 & 55.14\,/\,56.74 \\

InternVL3.5 + MaxInfo & 8B & 16 & 58.41\,/\,61.03 & 68.78\,/\,\textbf{73.54} & \uline{74.42}\,/\,\uline{75.00} & 57.77\,/\,59.47 & 50.53\,/\,53.72 \\

InternVL3.5 + Q-Frame & 8B & 4+8+32 & 60.58\,/\,61.71 & \textbf{69.84}\,/\,\uline{72.49} & 73.84\,/\,\textbf{76.16} & 60.92\,/\,60.44 & 53.19\,/\,54.61 \\

InternVL3.5 + QaF + LongCLIP & 8B & 16 & \uline{61.48}\,/\,\uline{64.92} & 68.78\,/\,\textbf{73.54} & 72.67\,/\,\textbf{76.16} & 63.11\,/\,\uline{66.75} & 54.43\,/\,57.27 \\

InternVL3.5 + QAaF + LongCLIP & 8B & 16 & \textbf{63.20}\,/\,\textbf{66.57} & \uline{69.31}\,/\,\textbf{73.54} & 72.09\,/\,74.42 & \textbf{64.08}\,/\,\textbf{67.96} & \textbf{57.80}\,/\,\textbf{60.82} \\

\bottomrule
\end{tabular}%
}
\label{tab:LongVideoBench_results}
\end{table} 
Table~\ref{tab:LongVideoBench_results} shows that InternVL3.5 with QAaF + LongCLIP achieves the best overall performance both without and with subtitles, scoring 63.20 and 66.57, respectively. It improves upon the InternVL3.5 uniform baseline by 4.49 points without subtitles and 4.79 points with subtitles. QAaF also provides consistent gains for the other MLLMs: it improves Qwen2-VL from 54.10/55.80 to 58.80/60.20 and LLaVA-Video from 56.90/59.30 to 61.32/63.49. These results indicate that aligning frames with both the query and candidate answers is effective across different MLLM architectures. Subtitles further improve the overall QAaF performance by 1.40, 2.17, and 3.37 points for Qwen2-VL, LLaVA-Video, and InternVL3.5, respectively.

Performance varies across video durations. For videos of 8–15 seconds, LLaVA-Video with AKS obtains the best result without subtitles at 70.90, while several InternVL3.5 configurations achieve the best result with subtitles at 73.54. For videos of 15–60 seconds, InternVL3.5 with Uniform sampling performs best without subtitles at 76.16, while InternVL3.5 with Q-Frame or QaF reaches 76.16 with subtitles. On longer videos, the advantage of QAaF becomes clearer: InternVL3.5 with QAaF achieves the highest scores for both the 3–10 minute category, 64.08/67.96, and the 15–60 minute category, 57.80/60.82. Overall, QAaF is particularly effective for long-duration videos, where selecting answer-relevant frames can reduce the difficulty of representing extended video content under a fixed frame budget.


\section{Conclusion}
\label{sec:conclusion}

Five training-free, MLLM-agnostic, plug-and-play frame-selection methods for long-video understanding have been proposed within the past year. However, they have not yet been evaluated through a comprehensive head-to-head comparison. This paper fills that gap by evaluating the five methods using three MLLMs across three benchmarks. QAaF achieves the best
performance in 13 of the 15 aggregate evaluation settings and the second-best performance in the remaining two, while FOCUS
ranks second overall. Future work should include consistent ablation studies across all five methods.
\bibliographystyle{IEEEtran}
\bibliography{MLLMref}

@article{zhang2024llava-video,
  title={{LLaVA-Video}: Video instruction tuning with synthetic data},
  author={Zhang, Yuanhan and Wu, Jinming and Li, Wei and Li, Bo and Ma, Zejun and Liu, Ziwei and Li, Chunyuan},
  journal={arXiv preprint arXiv:2410.02713},
  year={2024}
}

@inproceedings{tang2025adaptive,
  title={Adaptive keyframe sampling for long video understanding},
  author={Tang, Xi and Qiu, Jihao and Xie, Lingxi and Tian, Yunjie and Jiao, Jianbin and Ye, Qixiang},
  booktitle={Proceedings of the Computer Vision and Pattern Recognition Conference},
  pages={29118--29128},
  year={2025}
}

@inproceedings{QframeZhang2025q,
  title={Q-frame: Query-aware frame selection and multi-resolution adaptation for video-llms},
  author={Zhang, Shaojie and Yang, Jiahui and Yin, Jianqin and Luo, Zhenbo and Luan, Jian},
  booktitle={Proceedings of the IEEE/CVF International Conference on Computer Vision},
  pages={22056--22065},
  year={2025}
}

@article{LLaVAMiniZhang2025llava,
  title={Llava-mini: Efficient image and video large multimodal models with one vision token},
  author={Zhang, Shaolei and Fang, Qingkai and Yang, Zhe and Feng, Yang},
  journal={arXiv preprint arXiv:2501.03895},
  year={2025}
}

@inproceedings{chen2024internvl,
  title={InternVL: Scaling up vision foundation models and aligning for generic visual-linguistic tasks},
  author={Zhe Chen and Jiannan Wu and Wenhai Wang and Weijie Su and Guo Chen and Sen Xing and Muyan Zhong and Qinglong Zhang and Xizhou Zhu and Lewei Lu and et al.},
  booktitle={Proceedings of the IEEE/CVF Conference on Computer Vision and Pattern Recognition},
  pages={24185--24198},
  year={2024}
}

@article{VideoLLaMA2Cheng2024videollama2,
  title={{VideoLLaMA 2}: Advancing spatial-temporal modeling and audio understanding in video-LLMs},
  author={Zesen Cheng and Sicong Leng and Hang Zhang and Yifei Xin and Xin Li and Guanzheng Chen and Yongxin Zhu and Wenqi Zhang and Ziyang Luo and Deli Zhao and et al.},
  journal={arXiv preprint arXiv:2406.07476},
  year={2024}
}

@article{jang2016gumbel_softmax,
  title={Categorical reparameterization with Gumbel-Softmax},
  author={Eric Jang and Shixiang Gu and Ben Poole},
  journal={arXiv preprint arXiv:1611.01144},
  year={2016}
}

@article{liang2024keyvideollm,
  title={KeyVideoLLM: Towards large-scale video keyframe selection},
  author={Hao Liang and Jiapeng Li and Tianyi Bai and Xijie Huang and Linzhuang Sun and Zhengren Wang and Conghui He and Bin Cui and Chong Chen and Wentao Zhang},
  journal={arXiv preprint arXiv:2407.03104},
  year={2024}
}

@inproceedings{lin2024vila,
  title={ViLA: On pre-training for visual language models},
  author={Ji Lin and Hongxu Yin and Wei Ping and Pavlo Molchanov and Mohammad Shoeybi and Song Han},
  booktitle={Proceedings of the IEEE/CVF Conference on Computer Vision and Pattern Recognition},
  pages={26689--26699},
  year={2024}
}

@inproceedings{video_chatgptMaaz2024,
  title={Video-ChatGPT: Towards detailed video understanding via large vision and language models},
  author={Muhammad Maaz and Hanoona Rasheed and Salman Khan and Fahad Shahbaz Khan},
  booktitle={Proceedings of the Annual Meeting of the Association for Computational Linguistics},
  year={2024}
}

@inproceedings{radford2021clip,
  title={Learning transferable visual models from natural language supervision},
  author={Alec Radford and Jong Wook Kim and Chris Hallacy and Aditya Ramesh and Gabriel Goh and Sandhini Agarwal and Girish Sastry and Amanda Askell and Pamela Mishkin and Jack Clark and et al.},
  booktitle={Proceedings of the International Conference on Machine Learning},
  pages={8748--8763},
  publisher={PMLR},
  year={2021}
}

@article{Qwen2VLwang2024qwen2_vl,
  title={{Qwen2-VL}: Enhancing vision-language model’s perception of the world at any resolution},
  author={Peng Wang and Shuai Bai and Sinan Tan and Shijie Wang and Zhihao Fan and Jinze Bai and Keqin Chen and Xuejing Liu and Jialin Wang and Wenbin Ge and et al.},
  journal={arXiv preprint arXiv:2409.12191},
  year={2024}
}

@article{lin2023_video_llava,
  title={Video-llava: Learning united visual representation by alignment before projection},
  author={Bin Lin and Yang Ye and Bin Zhu and Jiaxi Cui and Munan Ning and Peng Jin and Li Yuan},
  year={2023},
  note={\url{https://arxiv.org/abs/2311.10122}}
}

@article{lin2023_sphinx,
  title={Sphinx: The joint mixing of weights, tasks, and visual embeddings for multi-modal large language models},
  author={Ziyi Lin and Chris Liu and Renrui Zhang and Peng Gao and Longtian Qiu and Han Xiao and Han Qiu and Chen Lin and Wenqi Shao and Keqin Chen and Jiaming Han and Siyuan Huang and Yichi Zhang and Xuming He and Hongsheng Li and Yu Qiao},
  year={2023},
  note={\url{https://arxiv.org/abs/2311.07575}}
}

@misc{wang2022_internvideo,
  title={Internvideo: General video foundation models via generative and discriminative learning},
  author={Wang, Yi and Li, Kunchang and Li, Yizhuo and He, Yinan and Huang, Bingkun and Zhao, Zhiyu and Zhang, Hongjie and Xu, Jilan and Liu, Yi and Wang, Zun and Xing, Sen and Chen, Guo and Pan, Junting and Yu, Jiashuo and Wang, Yali and Wang, Limin and Qiao, Yu},
  year={2022},
  url={https://arxiv.org/abs/2212.03191}
}

@misc{uncompressedWideo2024,
  author       = "{Wikipedia contributors}",
  title        = "Uncompressed video",
  year         = "2024",
  howpublished = "\url{https://en.wikipedia.org/wiki/Uncompressed_video}",
  note         = "Accessed: 2026-04-10"
}

@article{wang2023vaquita,
  title={Vaquita: Enhancing alignment in llm-assisted video understanding},
  author={Wang, Yizhou and Zhang, Ruiyi and Wang, Haoliang and Bhattacharya, Uttaran and Fu, Yun and Wu, Gang},
  journal={arXiv preprint arXiv:2312.02310},
  year={2023}
}

@inproceedings{zhu2026focus,
  title={{FOCUS}: Efficient Keyframe Selection for Long Video Understanding},
  author={Zirui Zhu and H Xu and Y Luo and Y Liu and K Sarkar and Z Yang and Y You},
  booktitle={International Conference on Learning Representations (ICLR)},
  year={2026},
  url={https://openreview.net/forum?id=1OQKqLFcbB}
}

@inproceedings{hu2025mllm,
  title={M-llm based video frame selection for efficient video understanding},
  author={Hu, Kai and Gao, Feng and Nie, Xiaohan and Zhou, Peng and Tran, Son and Neiman, Tal and Wang, Lingyun and Shah, Mubarak and Hamid, Raffay and Yin, Bing and others},
  booktitle={Proceedings of the Computer Vision and Pattern Recognition Conference},
  pages={13702--13712},
  year={2025}
}

@inproceedings{li2026maxinfo,
  title={Maxinfo: A training-free key-frame selection method using maximum volume for enhanced video understanding},
  author={Li, Pengyi and Abdullaeva, Irina and Gambashidze, Alexander and Kuznetsov, Andrey and Oseledets, Ivan},
  booktitle={Proceedings of the IEEE/CVF Winter Conference on Applications of Computer Vision},
  pages={7198--7207},
  year={2026}
}

@misc{Islam2026QueryAlignedVFS,
  title={Query-Aligned Video Frame Selection for Long Video Understanding},
  author={Md. Safayet Islam and Dilip Sarkar and Liang Liang},
  year={2026},
  howpublished={Tech. Report},
  note = {will be uploaded to arXiv soon}
}

@book{poynton2012BookVideoCompression,
  author    = {Charles Poynton},
  title     = {Digital Video and HD: Algorithms and Interfaces},
  edition   = {2nd},
  publisher = {Morgan Kaufmann},
  year      = {2012},
  isbn      = {9780123919267}
}

@incollection{shapley1953value,
  author    = {Lloyd S. Shapley},
  title     = {A Value for n-Person Games},
  booktitle = {Contributions to the Theory of Games II},
  editor    = {Harold W. Kuhn and Albert W. Tucker},
  series    = {Annals of Mathematical Studies},
  volume    = {28},
  pages     = {307--317},
  publisher = {Princeton University Press},
  address   = {Princeton, NJ},
  year      = {1953}
}

@article{mikhalev2018rectangulaMaxVol,
  title={Rectangular maximum-volume submatrices and their applications},
  author={Mikhalev, Aleksandr and Oseledets, Ivan V},
  journal={Linear Algebra and its Applications},
  volume={538},
  pages={187--211},
  year={2018},
  publisher={Elsevier}
}
\end{document}